\documentclass[sigconf,nonacm]{acmart}

\usepackage{graphicx}
\usepackage{balance}
\usepackage{placeins}
\usepackage{array}

\renewcommand\footnotetextcopyrightpermission[1]{}
\setcopyright{none}

\title{Alignment in Time: Peak-Aware Orchestration for Long-Horizon Agentic Systems}

\author{Hanjing Shi}
\email{hasa23@lehigh.edu}
\affiliation{
  \institution{Lehigh University}
  \city{Bethlehem}
  \state{Pennsylvania}
  \country{USA}
}

\author{Dominic DiFranzo}
\email{djd219@lehigh.edu}
\affiliation{
  \institution{Lehigh University}
  \city{Bethlehem}
  \state{Pennsylvania}
  \country{USA}
}

\begin{document}

\begin{abstract}
Traditional AI alignment primarily focuses on individual model outputs; however, autonomous agents in long-horizon workflows require sustained reliability across entire interaction trajectories. We introduce \textbf{APEMO} (Affect-aware Peak-End Modulation for Orchestration), a runtime scheduling layer that optimizes computational allocation under fixed budgets by operationalizing temporal-affective signals. Instead of modifying model weights, APEMO detects trajectory instability through behavioral proxies and targets repairs at critical segments, such as peak moments and endings. Evaluation across multi-agent simulations and LLM-based planner--executor flows demonstrates that APEMO consistently enhances trajectory-level quality and reuse probability over structural orchestrators. Our results reframe alignment as a temporal control problem, offering a resilient engineering pathway for the development of long-horizon agentic systems.
\end{abstract}

\maketitle

\section{Introduction}

Artificial intelligence systems are rapidly transitioning from single-turn instruction followers to autonomous, multi-step agents operating within long-horizon workflows. Recent work on autonomous LLM agents has demonstrated sustained multi-step planning and open-ended interaction \cite{wang2023voyager}. In such environments, large language models (LLMs) plan, execute, and coordinate across specialized roles to solve open-ended tasks. Alignment in these settings is no longer solely a property of model parameters; it unfolds across time. System behavior accumulates over turns, and user expectations adapt throughout the interaction. Over extended trajectories, accumulated inconsistencies may lead to cumulative frustration or degradation of behavioral coherence.

Recent advances in AI alignment have focused on value specification and instruction tuning, particularly through reinforcement learning from human feedback (RLHF) \cite{christiano2017deep, ouyang2022training}. In parallel, the HCI community has established principles for transparency, trust calibration, and effective human--AI collaboration \cite{amershi2019guidelines, bansal2021does}. These approaches primarily operate at the model or interface level. They improve output correctness, explanation quality, and interaction clarity. However, they do not explicitly address how alignment unfolds across extended interaction sequences.

At the orchestration level, multi-agent and workflow frameworks have advanced structural optimization. Structured reasoning paradigms such as ReAct \cite{yao2023react} and automated workflow generation systems such as AFlow \cite{zhang2024aflow} focus on improving task accuracy, routing strategies, and computational efficiency. Yet these systems typically aggregate performance across steps. They rarely treat the temporal structure of interaction as an alignment variable. In long-horizon workflows, perceived reliability depends not only on average step correctness, but also on how systems recover from intermediate failures and how trajectories conclude.

Psychological research demonstrates that retrospective evaluations of experiences are temporally structured. The peak--end rule shows that judgments are disproportionately influenced by the most intense moment and the ending of an experience \cite{kahneman1993when, fredrickson1993duration}. This temporal asymmetry suggests that interaction quality cannot be fully captured by mean performance alone. If human evaluation is temporally weighted, then alignment mechanisms that ignore trajectory structure may systematically misalign with how users assess system reliability and reuse potential.

In this paper, we introduce \textbf{APEMO} (Affect-aware Peak-End Modulation for Orchestration), a temporal-affective orchestration layer for AI workflows. APEMO does not modify model weights or training objectives. Instead, it reallocates reasoning effort and repair across a trajectory under fixed computational budgets. We formulate temporal orchestration as a constrained multi-objective optimization problem that balances trajectory-level quality, reuse propensity, frustration-related behavioral signals, and coordination cost.

Through systematic evaluation across long-horizon single-agent trajectories and multi-agent workflows, we investigate whether temporal control can function as a bidirectional alignment layer. Rather than aligning only token-level outputs to predefined values, our approach aligns interaction dynamics with evaluation-sensitive trajectory properties. This reframes alignment as a property of entire interaction sequences rather than isolated predictions.

\begin{figure*}[t]
\centering
\includegraphics[width=\linewidth]{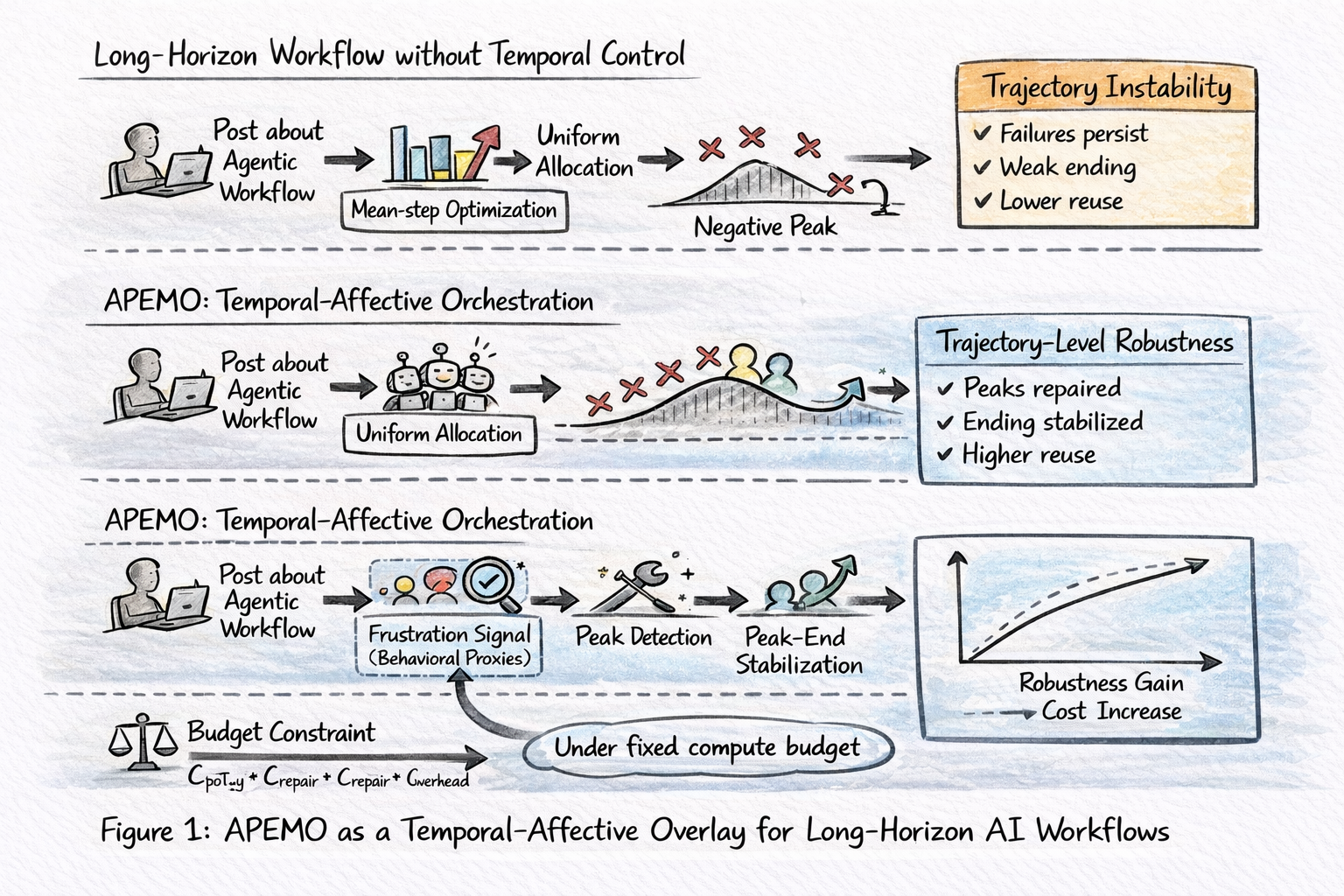}
\caption{
APEMO as a temporal-affective overlay for long-horizon workflows. 
Uniform allocation optimizes mean-step performance but allows negative peaks to persist. 
APEMO reallocates effort toward peak repair and ending stabilization under fixed compute budgets, 
improving trajectory-level robustness.
}
\label{fig:apemo_overview}
\end{figure*}

Figure~\ref{fig:apemo_overview} illustrates the core intuition of our approach. 
Under uniform budget allocation, negative peaks persist and endings remain weak, 
leading to trajectory-level instability. APEMO introduces temporal control that 
detects and repairs negative peaks while preserving a fixed compute budget.

\section{Related Work}

\subsection{Model Alignment and Preference Modeling}

Reinforcement Learning from Human Feedback (RLHF) aligns model outputs with human preferences by learning reward functions from labeled comparisons \cite{christiano2017deep, ouyang2022training}. However, human preference judgments are not temporally neutral. Prior work has shown that trajectory-level evaluations often exhibit non-Markovian structures, where judgments depend disproportionately on salient segments rather than average reward \cite{neurips2022nonmarkovian}. 

Most alignment methods treat such temporal weighting as bias to be corrected within reward modeling. In contrast, our work treats temporal asymmetry as a structural signal. Rather than modifying the reward function or retraining the model, we operationalize evaluation-sensitive weighting at runtime. This shifts alignment from a parameter-level optimization problem to a trajectory-level control problem.

Recent work on self-reflection and iterative reasoning, such as Tree-of-Thoughts \cite{yao2023tree}, Reflexion \cite{shinn2023reflexion}, and Self-Refine \cite{madaan2023selfrefine}, improves performance by expanding search depth or introducing additional reasoning passes. These approaches typically increase computational effort to enhance solution quality. In contrast, APEMO operates under a fixed global budget and reallocates compute across turns. Our objective is not to increase total reasoning depth, but to redistribute effort toward evaluation-critical segments such as negative peaks and final states.

\subsection{Agentic Orchestration and Multi-Agent Workflows}

The field has rapidly progressed from single-step inference to compound AI systems composed of planners, executors, critics, and memory modules. ReAct interleaves reasoning and acting to improve task performance \cite{yao2023react}. AFlow automates agentic workflow generation using structured search \cite{zhang2024aflow}. AgentOrchestra introduces hierarchical coordination via the Tool--Environment--Agent protocol \cite{agentorchestra2025}. More recent systems such as CORAL \cite{coral2601}, Kairos \cite{kairos2508}, and Orchestrator \cite{orchestrator2509} address information flow, latency, and long-horizon coordination efficiency.

These systems focus primarily on structural optimization, search efficiency, routing logic, or serving scalability. Their objectives typically aggregate performance across steps, implicitly approximating mean-step optimization. While some incorporate reflection or memory mechanisms, they rarely formalize temporal asymmetry as an optimization target. In particular, coordination cost is often treated as engineering overhead rather than an explicit trade-off variable.

Our work complements these orchestration architectures by introducing a temporal-affective overlay that operates orthogonally to workflow topology. Instead of redesigning agent roles or adding new modules, APEMO reallocates existing compute to stabilize evaluation-critical segments under a fixed budget. This distinguishes compute reallocation from compute expansion and frames coordination tax as a first-class variable in alignment analysis.

\subsection{Temporal Evaluation and Robust Interaction Trajectories}

The peak--end rule demonstrates that retrospective evaluations are disproportionately influenced by the most intense moment and the ending of an experience \cite{kahneman1993when, fredrickson1993duration}. This temporal asymmetry suggests that interaction quality cannot be fully captured by mean performance alone. If human assessment is temporally weighted, then alignment mechanisms that optimize uniform step performance may systematically diverge from user-perceived reliability.

In long-horizon AI systems, extended interaction introduces additional challenges, including behavioral drift and coherence degradation over time. Open-ended agents such as Voyager illustrate how multi-step autonomy increases exposure to trajectory instability \cite{wang2023voyager}. Memory architectures such as BMAM provide mechanisms for salience-aware retrieval \cite{bmam2026}, yet memory storage alone does not specify how compute should be redistributed when instability arises.

APEMO integrates temporal evaluation principles with runtime control by treating frustration-related behavioral proxies as actionable signals. Rather than enforcing uniform quality across all turns, it prioritizes negative-peak recovery and end-state stabilization under fixed computational constraints. 

To validate this control principle across abstraction levels, we adopt a stacked evidence strategy spanning Agent-Based Modeling (ABM), single-agent LLM reasoning, and multi-agent workflows. ABM isolates temporal dynamics under controlled Monte Carlo sampling; single-agent LLM experiments test robustness under neural stochasticity; and multi-agent evaluations verify that temporal control remains orthogonal to role decomposition. Together, these layers demonstrate that peak-aware compute reallocation is not a simulation artifact but a transferable orchestration principle.

\section{Methodology}

\subsection{Problem Formulation}

We model a long-horizon AI workflow as a trajectory of $T$ sequential turns. 
At each turn $t$, the system produces an intermediate output $y_t$ and generates observable behavioral signals. 
Rather than optimizing isolated step accuracy, we treat alignment as a trajectory-level property.

APEMO defines a runtime policy $\pi$ that maximizes the following constrained objective:

\begin{equation}
\max_{\pi} \mathbb{E}\left[\alpha Q + \beta R - \gamma F - \lambda C \right]
\end{equation}

where:

\begin{itemize}
    \item $Q$ denotes peak-end weighted trajectory quality,
    \item $R$ denotes reuse-related robustness,
    \item $F$ denotes cumulative frustration burden,
    \item $C$ denotes coordination cost.
\end{itemize}

All policies operate under identical model weights, decoding parameters, workflow topology, and shared computational caps. Realized coordination cost may differ by policy and is explicitly measured. 
APEMO differs from baselines only in runtime temporal allocation and repair scheduling.

Trajectory-level quality $Q$ is defined using peak-end weighting:

\begin{equation}
Q = w_p \max(q_{1\dots T}) + w_e \cdot \text{mean}(q_{T-1}, q_T)
\end{equation}

Coordination cost accumulates as:

\begin{equation}
C = C_{policy} + C_{repair} + C_{overhead}
\end{equation}

with the global constraint:

\[
C \le C_{max}
\]

This formulation reframes alignment as trajectory dynamics optimization rather than token-level correctness.

\subsection{Runtime Control Mechanism}

\begin{figure*}[t]
\centering
\includegraphics[width=\linewidth]{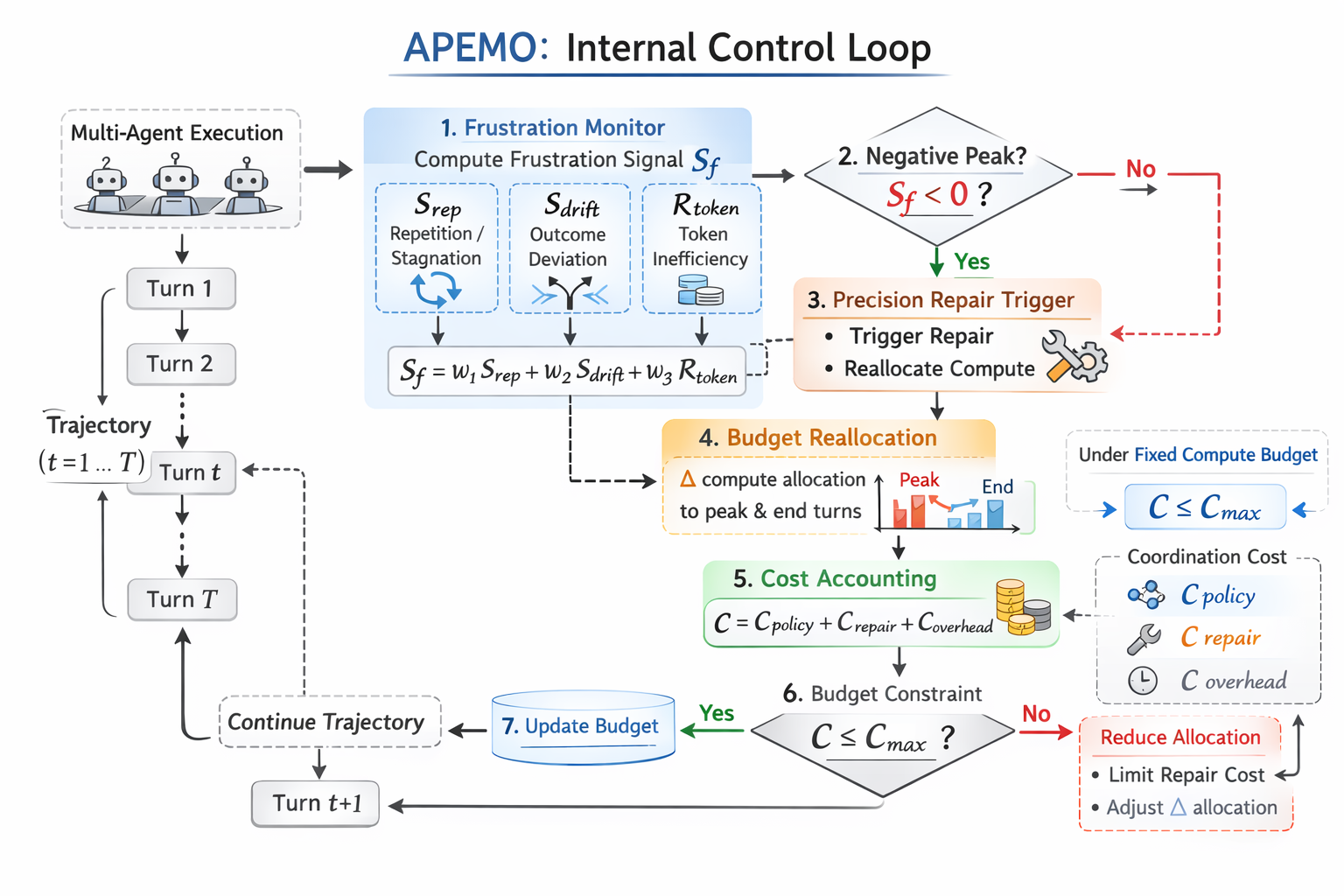}
\caption{
APEMO internal control loop. 
At each turn $t$, a frustration signal $S_f(t)$ is computed from behavioral proxies. 
If a negative peak is detected, a precision repair module reallocates compute 
toward peak and ending turns. 
Coordination cost accumulates under a fixed budget constraint $C \le C_{max}$.
}
\label{fig:apemo_internal}
\end{figure*}

Figure~\ref{fig:apemo_internal} illustrates the APEMO internal control loop.

At each turn, behavioral proxy signals are aggregated into a scalar frustration score $S_f(t)$. 
A peak-detection condition evaluates whether trajectory quality has entered a negative regime. 
When triggered, a precision repair operation reallocates compute from low-impact turns to peak or terminal segments.

The repair process does not modify model weights or workflow topology. 
Instead, it redistributes inference precision within the same global compute budget. 
Budget enforcement ensures that total coordination cost remains bounded.

Unlike topology-based orchestrators, APEMO operates as a temporal overlay that adjusts when and where reasoning precision is applied.

\subsection{Trade-off and Frontier Analysis}

Because temporal scheduling introduces coordination overhead, 
we analyze the trade-off between robustness gains and coordination cost.

Configurations are projected onto a Pareto frontier defined over $(R, C)$. 
A configuration is considered economically viable if improvements in reuse-related robustness exceed marginal coordination overhead.

This analysis isolates whether trajectory-level stabilization justifies additional orchestration cost under fixed budgets.

\section{Experimental Setup}

We evaluate APEMO under controlled equal-budget conditions to isolate the effect of temporal-affective orchestration. 
Our objective is not to optimize model training but to test whether runtime temporal control improves trajectory-level robustness.

All experiments are conducted using locally deployed small-model LLM families, including \texttt{llama3.2:1b}, \texttt{qwen2.5:1.5b}, and \texttt{gemma2:2b}, executed through an Ollama-based runtime. 
Prompt templates, decoding parameters, and sampling temperature are held constant across all conditions. 
No fine-tuning, reward modification, or architectural changes are applied.

Long-horizon evaluation uses $T=8$ turns to induce trajectory drift and potential negative peaks. 
Short-horizon evaluation uses $T=2$ to test whether improvements depend on extended context accumulation. 
Multi-agent settings adopt a fixed Planner--Executor--Critic topology across all variants, with APEMO applied solely as a temporal overlay.

Baselines include uniform budget allocation (mean-step optimization), structured reflective orchestration (\textit{plan\_execute} and \textit{plan\_execute\_reflect}), and role-based multi-agent flows without temporal reallocation (\textit{flow\_plain} and \textit{flow\_temporal}). 
All policies are evaluated under the same runtime protocol and global budget constraints; realized token and orchestration costs are then compared directly.

To evaluate recovery robustness, we introduce a controlled perturbation benchmark that injects mid-trajectory reasoning degradation to simulate negative peaks. 
Recovery quality is measured as post-perturbation stabilization relative to pre-trap performance.

Performance is evaluated using four trajectory-level metrics: peak-end weighted quality $Q$, reuse-related robustness $R$, cumulative frustration $F$, and coordination cost $C$.

We conduct both agent-based model (ABM) simulations and real LLM executions. 
ABM simulations are used for mechanism calibration and stress testing, while small-model LLM experiments provide the primary quantitative evidence reported in the main Results section. 
This separation avoids over-claiming from simulator-only dynamics while preserving a controlled mechanism-development layer.

\subsection{Primary Block Configuration Overview}

Table~\ref{tab:block_config} summarizes the main LLM blocks reported in Results, including run count and validity gate.

\begin{table*}[t]
\centering
\caption{Primary LLM block configuration summary used in main Results.}
\label{tab:block_config}
\setlength{\tabcolsep}{3pt}
\renewcommand{\arraystretch}{0.92}
\footnotesize
\resizebox{\textwidth}{!}{%
\begin{tabular}{lcccccc}
\hline
\textbf{Block} & \textbf{Models} & \textbf{Turns} & \textbf{Episodes} & \textbf{Runs (n)} & \textbf{Policies} & \textbf{no\_fallback\_rate} \\
\hline
Strict Long-Horizon & qwen2.5:1.5b, gemma2:2b & 8 & 2 & 20 & task\_affect, task\_peak\_end, apemo & 1.0 \\
Short-Horizon Boundary & llama3.2:1b, qwen2.5:1.5b, gemma2:2b & 2 & 2 & 21 & task\_affect, task\_peak\_end, apemo & 1.0 \\
Trap Recovery (strict) & gemma2:2b & 8 & 1 & 20 & task\_peak\_end, apemo & 1.0 \\
Multi-Agent Flow & qwen2.5:1.5b, gemma2:2b & 8 & 1 & 16 & flow\_plain, flow\_temporal, apemo & 1.0 \\
L3 Plan-Execute-Class & llama3.2:1b, qwen2.5:1.5b, gemma2:2b & 2 & 2 & 30 & plan\_execute, plan\_execute\_reflect, task\_peak\_end, apemo & 1.0 \\
L3 Plan-Execute-Class (Long-Horizon) & llama3.2:1b, qwen2.5:1.5b, gemma2:2b & 8 & 1 & 18 & plan\_execute, plan\_execute\_reflect, task\_peak\_end, apemo & 1.0 \\
\hline
\end{tabular}%
}%
\normalsize
\end{table*}

\subsection{Statistical Analysis}

All statistics are computed at the run level. 
Each run corresponds to a fixed configuration defined by model, seed, policy, and trajectory length. 
Episode-level metrics are aggregated within run to avoid pseudo-replication.

For each metric, we report mean, standard deviation (std), and standard error (se). 
Uncertainty is estimated using nonparametric bootstrap confidence intervals with 95\% coverage via resampling over runs.

Pairwise comparisons are expressed as delta metrics:

\[
\Delta = \text{Metric}_{\text{APEMO}} - \text{Metric}_{\text{Baseline}}.
\]

For each delta, we report the 95\% bootstrap confidence interval.

As a conservative small-sample robustness check, we additionally report two-sided sign-test p-values based on directional wins across runs. Key p-values are reported in the main Results text and full p-value tables are provided in block-level comparison files.
This nonparametric test avoids distributional assumptions.

Interpretation is gated by reliability validity. 
For each block, we compute a \textit{no\_fallback\_rate}. 
Blocks with incomplete validity are treated as directional evidence rather than primary statistical support.

All comparisons are theory-driven and restricted to predefined trajectory-level objectives. 
Primary endpoints were fixed before large-$n$ reruns. 
These include mean quality, reuse probability, reuse-per-cost, average frustration, 
and trap endpoint and rebound metrics.
We emphasize effect sizes and uncertainty intervals over binary significance claims.

\section{Results}

We report results with a staged validation strategy progressing from controlled trajectory tests to perturbation recovery and multi-agent extensions. All statistics are computed at the run level. Reported intervals are nonparametric 95\% bootstrap confidence intervals.

\subsection{Long-Horizon Trajectory Performance}

We first evaluate APEMO under strict long-horizon settings ($T=8$) with cross-model runs and $n=20$ independent configurations. All runs completed without fallback (no\_fallback\_rate = 1.0).

Relative to the peak-end baseline, APEMO improves mean trajectory quality by $+0.0791$ (95\% CI [0.0525, 0.1055], sign-test $p=4.01\times10^{-5}$) and reuse probability by $+0.0609$ (95\% CI [0.0383, 0.0826], $p=4.01\times10^{-5}$). Average frustration decreases by $-0.0172$ (95\% CI [-0.0306, -0.0071], $p=0.115$).

Relative to the affect baseline, gains are larger: quality $+0.1874$ (95\% CI [0.1596, 0.2159], $p=1.91\times10^{-6}$), reuse probability $+0.1161$ (95\% CI [0.0861, 0.1433], $p=4.01\times10^{-5}$), and frustration $-0.0543$ (95\% CI [-0.0805, -0.0356], $p=1.91\times10^{-6}$).

Table~\ref{tab:strict_long_revised} summarizes the primary long-horizon results.

\begin{table}[t]
\footnotesize 
\centering
\caption{Strict long-horizon evaluation ($T=8$, $n=20$). Values are mean deltas with 95\% bootstrap CIs.}
\label{tab:strict_long_revised}
\setlength{\tabcolsep}{6pt} 
\begin{tabular}{lccc}
\hline
\textbf{Baseline} & \textbf{\shortstack{Mean\\Quality}} & \textbf{\shortstack{Reuse\\Prob.}} & \textbf{\shortstack{Avg.\\Frust.}} \\
\hline
\addlinespace[5pt]
task\_peak\_end
& \shortstack{$+0.0791$\\ \tiny{[0.0525, 0.1055]}}
& \shortstack{$+0.0609$\\ \tiny{[0.0383, 0.0826]}}
& \shortstack{$-0.0172$\\ \tiny{[-0.0306, -0.0071]}} \\
\addlinespace[5pt] % 增加行间距，防止上下两行太挤
task\_affect
& \shortstack{$+0.1874$\\ \tiny{[0.1596, 0.2159]}}
& \shortstack{$+0.1161$\\ \tiny{[0.0861, 0.1433]}}
& \shortstack{$-0.0543$\\ \tiny{[-0.0805, -0.0356]}} \\
\hline
\end{tabular}
\end{table}

\begin{figure*}[t]
\centering
\includegraphics[width=\textwidth]{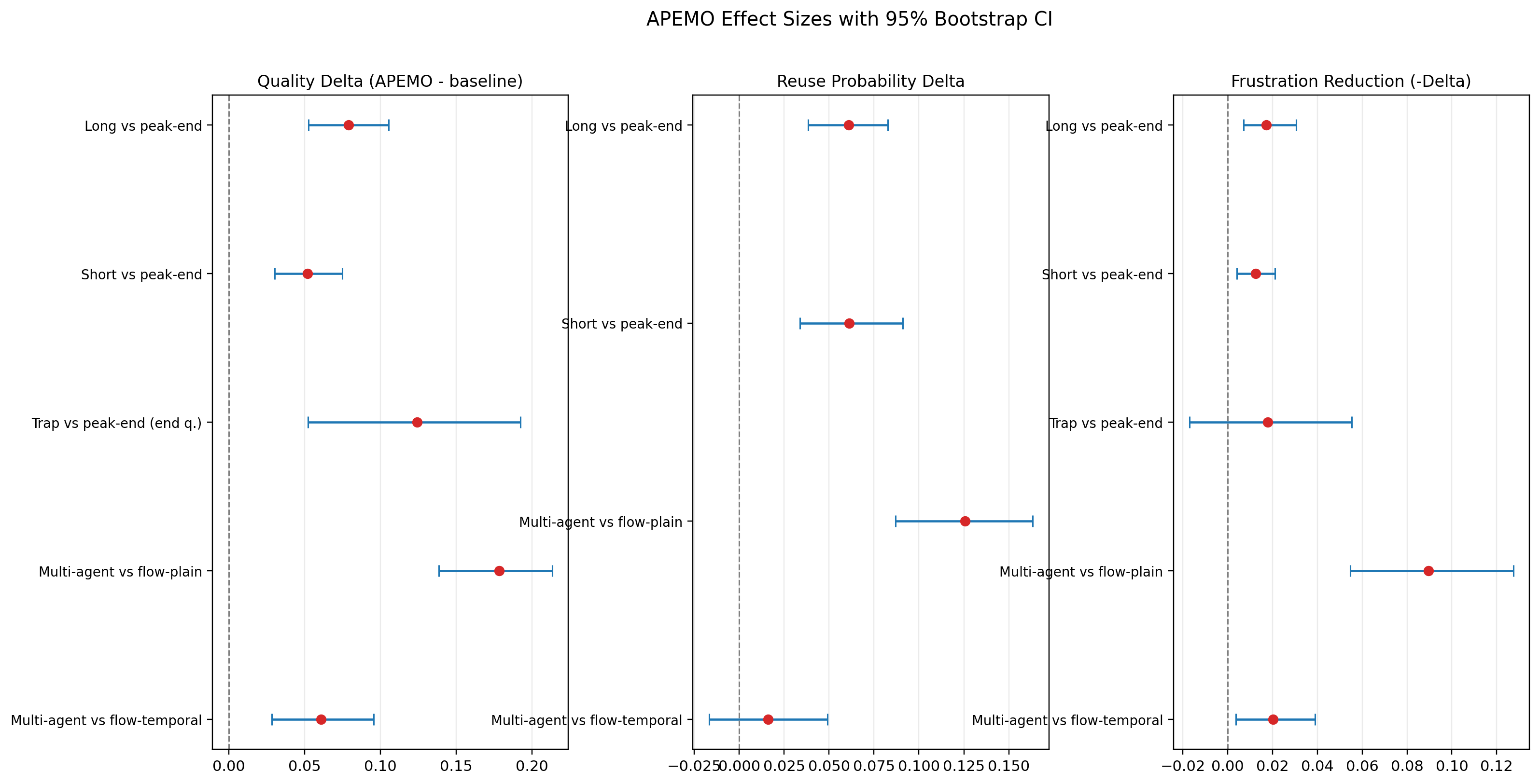}
\caption{Forest-style effect plot for key deltas (APEMO minus baseline) with 95\% bootstrap CIs.}
\label{fig:forest_effects}
\end{figure*}

Figure~\ref{fig:forest_effects} reports effect sizes and confidence intervals across long-horizon, short-horizon, trap, and multi-agent blocks.

\subsection{ABM Mechanism Consistency (Support Layer)}

To verify mechanism consistency beyond specific model families, we run high-volume ABM stress tests under matched policy structure. ABM results are directionally aligned with LLM findings (better endpoint/recovery behavior with lower frustration), supporting the control mechanism itself rather than any single LLM idiosyncrasy.

Because simulator effect magnitudes can be inflated relative to real LLM stochastic regimes, we treat ABM as support-layer evidence and report full ABM effect tables in supplementary material rather than using them as the primary basis of claims.

\subsection{Short-Horizon Regime Boundary}

To test horizon sensitivity, we evaluate short-horizon runs ($T=2$, $n=21$, no\_fallback\_rate = $1.0$).

APEMO improves mean quality by $+0.0520$ (95\% CI [0.0303, 0.0750], $p=0.0072$) and reuse probability by $+0.0613$ (95\% CI [0.0339, 0.0910], $p=0.0072$) over task\_peak\_end. Reuse-per-cost remains positive ($+0.0051$, 95\% CI [0.0015, 0.0085], $p=0.0266$), though effect magnitudes are smaller than in long-horizon settings.

This boundary condition suggests that coordination overhead amortizes more effectively when trajectories are deeper, consistent with the trajectory-level control hypothesis developed in Discussion.

\subsection{Negative-Peak Perturbation and Recovery}

We introduce controlled negative-peak perturbations ($n=20$, no\_fallback\_rate = $1.0$).

APEMO improves endpoint delivery quality by $+0.1243$ (95\% CI [0.0522, 0.1924], $p=0.0118$) relative to task\_peak\_end. Rebound metrics remain directionally positive but variance-sensitive: trap\_quality\_rebound2 $+0.0807$ (95\% CI [0.0075, 0.1540], $p=0.824$), while trap\_frustration\_drop2 $+0.0439$ (95\% CI [-0.0084, 0.0943], $p=0.263$) includes zero. The CI and sign-test discrepancy on rebound is expected under this conservative test with mixed directional wins.

\begin{figure}[t]
\centering
\includegraphics[width=\columnwidth]{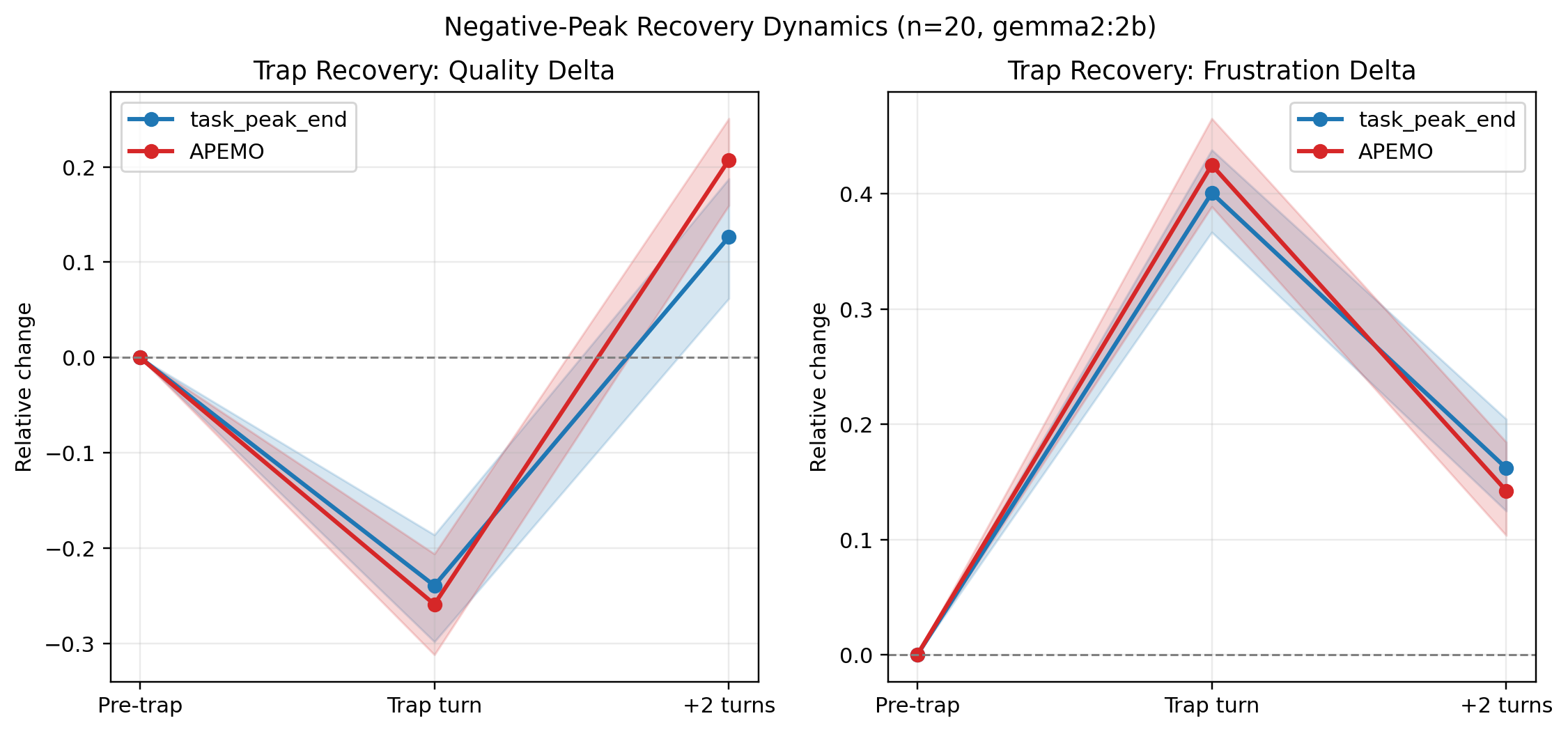}
\caption{Trap recovery dynamics in the $n=20$ LLM trap block, derived from episode-level trap drop/rebound metrics. Left: quality change around trap; right: frustration change around trap. Shaded areas are 95\% bootstrap CIs.}
\label{fig:trap_recovery}
\end{figure}

These results indicate that endpoint stabilization is robust, whereas intermediate rebound amplitudes depend more heavily on stochastic cross-model dynamics. Figure~\ref{fig:trap_recovery} shows trap-response dynamics. APEMO shows weaker quality collapse at the trap turn and improved endpoint recovery behavior.

\subsection{Multi-Agent Workflow Extension}

We extend APEMO above a Planner--Executor--Critic topology ($n=16$, no\_fallback\_rate = $1.0$).

Relative to plain multi-agent flow, APEMO improves quality by $+0.1782$ (95\% CI [0.1386, 0.2132], $p=3.05\times10^{-5}$) and reuse probability by $+0.1257$ (95\% CI [0.0869, 0.1633], $p=5.19\times10^{-4}$). Relative to the stronger temporal-flow baseline, quality remains positive ($+0.0609$, 95\% CI [0.0285, 0.0956], $p=0.0768$) while reuse and reuse-per-cost are near ties (confidence intervals include zero; $p=0.4545$ and $p=0.8036$).

This pattern suggests that temporal-affective scheduling complements structural role decomposition and yields its strongest improvements when no prior temporal control is present.

\subsection{Plan-Execute-Class Baseline Alignment (L3)}

We expanded L3 to a larger cross-model block with $n=30$ valid turn-2 runs. 
We compare plan\_execute, plan\_execute\_reflect, 
task\_peak\_end, and APEMO under the same runtime protocol.

Relative to \textit{plan\_execute}, APEMO improves quality by $+0.1028$ (95\% CI [0.0843, 0.1218], $p=1.86\times10^{-9}$), reuse probability by $+0.0547$ (95\% CI [0.0381, 0.0724], $p=5.95\times10^{-5}$), and reduces frustration by $-0.0181$ (95\% CI [-0.0269, -0.0105], $p=3.25\times10^{-4}$). Reuse-per-cost is also positive ($+0.0027$, 95\% CI [0.0006, 0.0048], $p=0.0428$).

Relative to \textit{plan\_execute\_reflect}, APEMO remains positive on quality ($+0.0370$, 95\% CI [0.0229, 0.0521], $p=5.95\times10^{-5}$) and reuse probability ($+0.0211$, 95\% CI [0.0081, 0.0347], $p=0.0052$), while reuse-per-cost and frustration remain near ties (confidence intervals include zero).

To address long-horizon external-baseline alignment directly, we further run an L3 turn-8 cross-model block (\textit{n}=18, no\_fallback\_rate = 1.0). Relative to \textit{plan\_execute}, APEMO improves mean quality by $+0.1259$ (95\% CI [0.0987, 0.1521], $p=7.63\times10^{-6}$), reuse probability by $+0.1223$ (95\% CI [0.0857, 0.1573], $p=1.45\times10^{-4}$), and average frustration by $-0.0193$ (95\% CI [-0.0295, -0.0100], $p=0.0013$). Relative to \textit{plan\_execute\_reflect}, quality remains positive at $+0.1881$ (95\% CI [0.1557, 0.2247], $p=7.63\times10^{-6}$), with reuse and reuse-per-cost also positive.

Against \textit{task\_peak\_end} at turn-8, APEMO still improves quality ($+0.0988$, 95\% CI [0.0631, 0.1356], $p=1.45\times10^{-4}$) and frustration ($-0.0096$, 95\% CI [-0.0159, -0.0047], $p=0.0075$), while reuse and reuse-per-cost become near ties. This reinforces the earlier boundary: L3 supports directional orthogonality, but efficiency deltas remain baseline-sensitive.

\subsection{Coordination Frontier}

We quantify trade-offs between robustness and orchestration cost using relative quality gain and realized total coordination cost increase under shared runtime constraints.

\begin{table}[t]
\small
\centering
\caption{Coordination frontier summary (large-$n$ blocks). Percentages are relative deltas. All rows use mean-quality gain except Trap Endpoint, which uses endpoint-quality gain by design.}
\label{tab:frontier_revised}
\begin{tabular}{p{0.55\columnwidth}cc}
\hline
\textbf{Setting} & \textbf{Quality Gain} & \textbf{Cost Increase} \\
\hline
Long-Horizon ($T{=}8$, vs peak-end) & $+14.49\%$ & $+6.28\%$ \\
Long-Horizon ($T{=}8$, vs affect) & $+42.86\%$ & $+8.15\%$ \\
Short-Horizon ($T{=}2$, vs peak-end) & $+8.48\%$ & $+3.39\%$ \\
Trap Endpoint (vs peak-end) & $+29.90\%$ & $+5.33\%$ \\
Multi-Agent (vs flow\_plain) & $+41.25\%$ & $+7.14\%$ \\
\hline
\end{tabular}
\end{table}

\begin{figure}[t]
\centering
\includegraphics[width=\columnwidth]{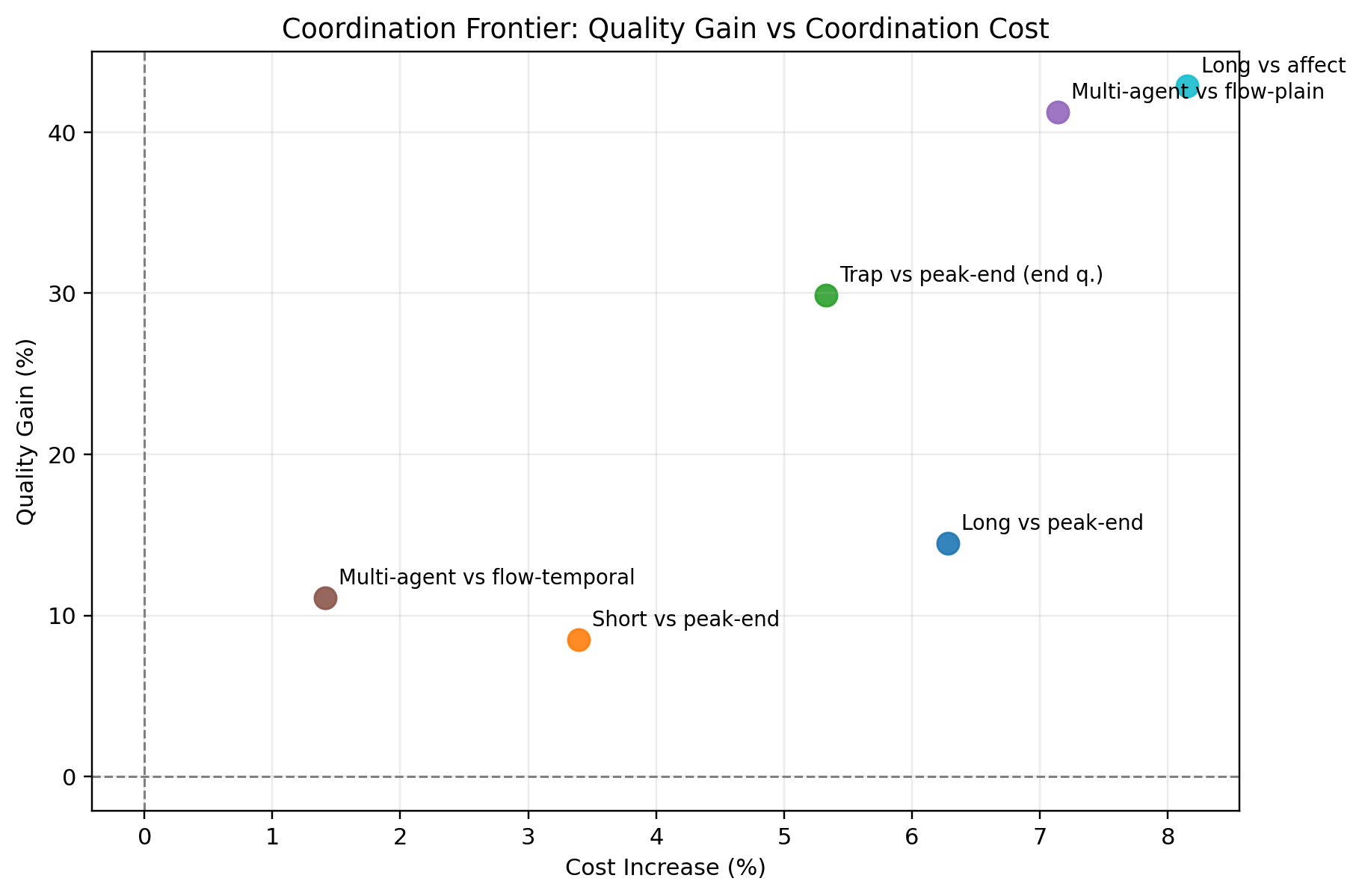}
\caption{Coordination frontier (quality gain vs cost increase). Each point is one large-$n$ block/comparison.}
\label{fig:frontier_scatter}
\end{figure}

Figure~\ref{fig:frontier_scatter} visualizes the frontier: Long-horizon and multi-agent plain-flow settings lie in a high-gain region where relative quality improvements dominate realized coordination cost increases. In short-horizon or strong-temporal baseline settings, gains remain positive but move toward diminishing-return regions.

\subsection{Scope and Boundary Conditions}

Cross-model trap evidence exhibits wider confidence intervals than strict non-trap long-horizon blocks, reflecting perturbation sensitivity and role-level interaction variance. We therefore treat cross-model trap runs as directional support and use the strict trap block (n=20, no\_fallback\_rate=1.0) as the primary perturbation evidence.

External-orchestrator alignment in this version targets plan-execute-class baselines. We now include both a larger short-horizon L3 block (turn-2, n=30) and a long-horizon L3 extension (turn-8, n=18), but we still do not claim full long-horizon parity against industrial workflow generators.

We therefore interpret APEMO as a trajectory-level orchestration layer that demonstrates stable benefits in long-horizon and plain-flow multi-agent regimes, with explicit boundary conditions in shallow trajectories, strong temporalized baselines, and currently limited long-horizon external-orchestrator comparisons.

\section{Discussion}

\subsection{Trajectory Alignment as a Control Problem}

Most contemporary alignment research treats alignment as a property of model outputs. Reward modeling and RLHF optimize for response-level preference consistency \cite{christiano2017deep, ouyang2022training}, while process supervision approaches refine reasoning traces at the step level \cite{lightman2023lets, uesato2022solving}. In parallel, HCI scholarship has emphasized trust calibration, transparency, and appropriate reliance in human–automation interaction \cite{amershi2019guidelines, lee2004trust, parasuraman1997humans}.

Our results suggest a complementary perspective: alignment in long-horizon workflows is not solely a property of outputs or reasoning chains, but of how interactions unfold across time. Psychological evidence demonstrates that retrospective evaluation is temporally asymmetric, disproportionately influenced by salient peaks and endings \cite{kahneman1993when, fredrickson1993duration}. Rather than treating this as bias to be corrected, APEMO treats temporal weighting as a predictable structural feature of evaluation.

This motivates a broader theoretical claim: many alignment pipelines implicitly assume a stationary mapping between local behavior and utility across a trajectory. Long-horizon interaction violates this assumption. Evaluation salience shifts over time as user state, task uncertainty, and accumulated context change, so identical local errors can have different global consequences depending on when they occur. In effect, alignment faces a within-trajectory temporal distribution shift, not only a dataset-level shift. Static reward alignment is therefore necessary but incomplete for agentic workflows. Temporal control policies are needed to track and correct this non-stationarity during execution.

Under shared runtime budget caps, reallocating effort across a trajectory can shift evaluation-sensitive outcomes without modifying token-level likelihoods or retraining reward models. In this sense, alignment becomes a runtime control problem over sequences. The locus of intervention moves from parameter space to temporal allocation.

\subsection{Orthogonality to Structural Agent Design}

Recent advances in agentic systems emphasize structural organization: reasoning–acting interleaving \cite{yao2023react}, planner–executor architectures \cite{zhang2024aflow}, hierarchical coordination \cite{agentorchestra2025}, or multi-agent conversational frameworks \cite{wu2023autogen, hong2024metagpt}. These systems restructure *who does what*.

Our findings indicate that temporal scheduling governs a different axis: *when and where compute is invested*. Even when planner–executor–critic roles remain unchanged, trajectory robustness increases under temporal-affective control. This suggests that structural orchestration and temporal allocation address separable dimensions of reliability.

This separation clarifies the engineering contribution of APEMO. It does not compete with topology search or role design; it overlays them. Structural frameworks define workflow composition, while APEMO governs evaluation-sensitive saliency within that composition.

\subsection{Mechanistic Interpretation and Stability Dynamics}

The empirical pattern across ABM, single-agent LLM, and multi-agent settings reveals two interacting stability dynamics.

Ending stabilization amplifies perceived reliability. Because final states disproportionately shape retrospective judgment \cite{kahneman1993when}, targeted compute allocation near termination improves reuse probability and evaluation outcomes even when intermediate steps are imperfect.

Early instability detection dampens cascading degradation. Behavioral proxies such as repetition similarity and context drift serve as low-cost indicators of trajectory destabilization. Precision repair prevents local perturbations from propagating across turns. This resembles adaptive compute principles in which effort is scaled based on contextual uncertainty rather than held constant \cite{schuster2022confident}.

The strength of long-horizon gains is therefore not accidental. In shallow trajectories, coordination overhead cannot amortize. In deeper trajectories, temporal control compounds across turns, producing superlinear robustness effects relative to uniform allocation.

\subsection{Implications for Human–AI Interaction}

Trust in automation is shaped not only by accuracy but by consistency and recovery behavior over time \cite{dzindolet2003automation, madhavan2007trust, hancock2023how}. Systems that degrade unpredictably across extended interactions undermine reliance even if average performance remains high.

APEMO suggests that reliability can be engineered at the temporal level. By shaping peaks and endings, systems can stabilize retrospective trust formation. This reframes alignment as a property of interaction trajectories rather than isolated answers. In long-running agentic workflows—such as open-ended embodied agents \cite{wang2023voyager} or tool-rich multi-agent systems—trajectory-level control may be critical for sustained collaboration.

\subsection{Limitations}

Several boundaries remain.

Trap-recovery micro-dynamics remain variance-sensitive in certain cross-model regimes. While endpoint stabilization is robust, rebound-style metrics fluctuate with model family and perturbation profile.

Comparisons against external orchestrators are restricted to plan–execute-style baselines rather than complete reproductions of industrial multi-agent systems \cite{zhang2024aflow, coral2601}. The results therefore demonstrate orthogonality rather than superiority over full-stack orchestration pipelines.

Human-subject validation is absent. Although trajectory-level metrics improve consistently, subjective trust, perceived competence, and reuse intent require direct empirical measurement \cite{lee2004trust, endsley1995out}.

Finally, real-world deployment may introduce orchestration overhead beyond the controlled cost accounting used here. Continuous monitoring and scheduling logic can introduce latency trade-offs in highly interactive settings. While coordination cost is explicitly modeled within fixed-budget constraints, end-to-end serving impact remains an open systems question \cite{kairos2508, kaplan2020scaling}.

\subsection{Future Work}

A natural extension is empirical validation of trajectory alignment in human studies. Measuring how peak-aware scheduling influences trust calibration, delegation decisions, and long-term collaboration stability would connect runtime control directly to human-centered automation theory \cite{parasuraman2000model, legler2025human}.

Another direction is adaptive scheduler learning. Rather than fixed thresholds, temporal allocation policies could be meta-learned across tasks and model families. Integrating runtime scheduling with process-supervision or reflective refinement loops \cite{shinn2023reflexion, madaan2023selfrefine} may yield systems that both allocate compute strategically and improve reasoning quality.

Finally, extending temporal control to frontier-scale models and longer contexts will clarify scaling properties. Larger models may exhibit intrinsic stabilization mechanisms that interact with scheduling policies. Understanding these interactions will determine whether trajectory alignment remains a critical layer as model capability increases.

\section{Conclusion}

This work introduces a temporal-affective perspective on alignment in long-horizon AI workflows. While prior approaches primarily optimize parameters, reward models, or structural agent topologies, we demonstrate that runtime temporal control constitutes an independent axis of reliability.

APEMO reallocates reasoning effort across interaction trajectories without modifying model weights or expanding model capacity. Through stacked validation across agent-based simulations, single-agent LLM execution, and multi-agent orchestration flows, we show that peak-aware scheduling improves trajectory-level robustness, stabilizes endpoint delivery, and increases reuse probability under shared runtime constraints, while explicitly quantifying realized coordination-cost differences.

These results suggest that alignment is not merely an outcome of model training or interface transparency. It is also shaped by how computation is distributed across time. Treating temporal saliency as a controllable signal opens a complementary pathway for engineering resilient agentic systems.

As AI systems increasingly operate over extended, autonomous workflows, trajectory-level control may become as critical as parameter level alignment. This work positions temporal orchestration as a principled and cost-aware mechanism for sustaining reliability in long-horizon interaction.

\bibliographystyle{ACM-Reference-Format}
\bibliography{references}

@inproceedings{christiano2017deep,
  title={Deep Reinforcement Learning from Human Preferences},
  author={Christiano, Paul F. and Leike, Jan and Brown, Tom and Martic, Miljan and Legg, Shane and Amodei, Dario},
  booktitle={Advances in Neural Information Processing Systems},
  year={2017}
}

@article{ouyang2022training,
  title={Training language models to follow instructions with human feedback},
  author={Ouyang, Long and Wu, Jeffrey and Jiang, Xu and Almeida, Diogo and Wainwright, Carroll and Mishkin, Pamela and Zhang, Chong and Agarwal, Sandhini and Slama, Katarina and Ray, Alex and others},
  journal={Advances in neural information processing systems},
  volume={35},
  pages={27730--27744},
  year={2022}
}

@inproceedings{amershi2019guidelines,
  title={Guidelines for human-AI interaction},
  author={Amershi, Saleema and Weld, Dan and Vorvoreanu, Mihaela and Fourney, Adam and Nushi, Besmira and Collisson, Penny and Suh, Jina and Iqbal, Shamsi and Bennett, Paul N and Inkpen, Kori and others},
  booktitle={Proceedings of the 2019 chi conference on human factors in computing systems},
  pages={1--13},
  year={2019}
}

@inproceedings{bansal2021does,
  title={Does the whole exceed its parts? the effect of ai explanations on complementary team performance},
  author={Bansal, Gagan and Wu, Tongshuang and Zhou, Joyce and Fok, Raymond and Nushi, Besmira and Kamar, Ece and Ribeiro, Marco Tulio and Weld, Daniel},
  booktitle={Proceedings of the 2021 CHI conference on human factors in computing systems},
  pages={1--16},
  year={2021}
}

@inproceedings{yao2023react,
  title={React: Synergizing reasoning and acting in language models},
  author={Yao, Shunyu and Zhao, Jeffrey and Yu, Dian and Du, Nan and Shafran, Izhak and Narasimhan, Karthik R and Cao, Yuan},
  booktitle={The eleventh international conference on learning representations},
  year={2022}
}

@article{zhang2024aflow,
  title={Aflow: Automating agentic workflow generation},
  author={Zhang, Jiayi and Xiang, Jinyu and Yu, Zhaoyang and Teng, Fengwei and Chen, Xionghui and Chen, Jiaqi and Zhuge, Mingchen and Cheng, Xin and Hong, Sirui and Wang, Jinlin and others},
  journal={arXiv preprint arXiv:2410.10762},
  year={2024}
}

@article{neurips2022nonmarkovian,
  title={Non-markovian reward modelling from trajectory labels via interpretable multiple instance learning},
  author={Early, Joseph and Bewley, Tom and Evers, Christine and Ramchurn, Sarvapali},
  journal={Advances in Neural Information Processing Systems},
  volume={35},
  pages={27652--27663},
  year={2022}
}

@article{agentorchestra2025,
  title={AgentOrchestra: Orchestrating hierarchical multi-agent intelligence with the Tool-Environment-Agent (TEA) protocol},
  author={Zhang, Wentao and Zeng, Liang and Xiao, Yuzhen and Li, Yongcong and Zhao, Yilei and Cui, Ce and Liu, Yang and An, Bo},
  year={2025}
}

@article{bmam2026,
  title={BMAM: Brain-inspired Multi-Agent Memory Framework},
  author={Li, Yang and Liu, Jiaxiang and Wang, Yusong and Wu, Yujie and Xu, Mingkun},
  journal={arXiv preprint arXiv:2601.20465},
  year={2026}
}

@article{kahneman1993when,
  title={When more pain is preferred to less: Adding a better end},
  author={Kahneman, Daniel and Fredrickson, Barbara L and Schreiber, Charles A and Redelmeier, Donald A},
  journal={Psychological science},
  volume={4},
  number={6},
  pages={401--405},
  year={1993},
  publisher={SAGE Publications Sage CA: Los Angeles, CA}
}

@article{fredrickson1993duration,
  title={Duration Neglect in Retrospective Evaluations of Affective Episodes},
  author={Fredrickson, Barbara L. and Kahneman, Daniel},
  journal={Journal of Personality and Social Psychology},
  volume={65},
  number={1},
  pages={45--55},
  year={1993}
}

@article{kairos2508,
  title={Kairos: Low-latency multi-agent serving with shared llms and excessive loads in the public cloud},
  author={Chen, Jinyuan and Shi, Jiuchen and Chen, Quan and Guo, Minyi},
  journal={arXiv preprint arXiv:2508.06948},
  year={2025}
}

@article{orchestrator2509,
  title={Orchestrator: Active Inference for Multi-Agent Systems in Long-Horizon Tasks},
  author={Beckenbauer, Lukas and Loewe, Johannes-Lucas and Zheng, Ge and Brintrup, Alexandra},
  journal={arXiv preprint arXiv:2509.05651},
  year={2025}
}

@article{coral2601,
  title={Beyond Rule-Based Workflows: An Information-Flow-Orchestrated Multi-Agents Paradigm via Agent-to-Agent Communication from CORAL},
  author={Ren, Xinxing and Zang, Quagmire and Forder, Caelum and Deb, Suman and Tahir, Ahsen and Georgio, Roman J and Carroll, Peter and Guo, Zekun},
  journal={arXiv preprint arXiv:2601.09883},
  year={2026}
}

@article{wang2023voyager,
  title={Voyager: An open-ended embodied agent with large language models},
  author={Wang, Guanzhi and Xie, Yuqi and Jiang, Yunfan and Mandlekar, Ajay and Xiao, Chaowei and Zhu, Yuke and Fan, Linxi and Anandkumar, Anima},
  journal={arXiv preprint arXiv:2305.16291},
  year={2023}
}

@article{yao2023tree,
  title={Tree of thoughts: Deliberate problem solving with large language models},
  author={Yao, Shunyu and Yu, Dian and Zhao, Jeffrey and Shafran, Izhak and Griffiths, Tom and Cao, Yuan and Narasimhan, Karthik},
  journal={Advances in neural information processing systems},
  volume={36},
  pages={11809--11822},
  year={2023}
}

@inproceedings{shinn2023reflexion,
  title={Reflexion: Language Agents with Verbal Reinforcement Learning},
  author={Shinn, Noah and others},
  booktitle={NeurIPS},
  year={2023}
}

@article{madaan2023selfrefine,
  title={Self-refine: Iterative refinement with self-feedback},
  author={Madaan, Aman and Tandon, Niket and Gupta, Prakhar and Hallinan, Skyler and Gao, Luyu and Wiegreffe, Sarah and Alon, Uri and Dziri, Nouha and Prabhumoye, Shrimai and Yang, Yiming and others},
  journal={Advances in neural information processing systems},
  volume={36},
  pages={46534--46594},
  year={2023}
}

@article{lee2004trust,
  title={Trust in Automation: Designing for Appropriate Reliance},
  author={Lee, John D. and See, Katrina A.},
  journal={Human Factors},
  volume={46},
  number={1},
  pages={50--80},
  year={2004},
  doi={10.1518/hfes.46.1.50_30392}
}

@article{parasuraman1997humans,
  title={Humans and Automation: Use, Misuse, Disuse, and Abuse},
  author={Parasuraman, Raja and Riley, Victor},
  journal={Human Factors},
  volume={39},
  number={2},
  pages={230--253},
  year={1997},
  doi={10.1518/001872097778543886}
}

@article{parasuraman2000model,
  title={A Model for Types and Levels of Human Interaction with Automation},
  author={Parasuraman, Raja and Sheridan, Thomas B. and Wickens, Christopher D.},
  journal={IEEE Transactions on Systems, Man, and Cybernetics - Part A: Systems and Humans},
  volume={30},
  number={3},
  pages={286--297},
  year={2000},
  doi={10.1109/3468.844354}
}

@article{endsley1995out,
  title={Toward a Theory of Situation Awareness in Dynamic Systems},
  author={Endsley, Mica R.},
  journal={Human Factors},
  volume={37},
  number={1},
  pages={32--64},
  year={1995},
  doi={10.1518/001872095779049543}
}

@article{dzindolet2003automation,
  title={The Role of Trust in Automation Reliance},
  author={Dzindolet, Mary T. and Peterson, Scott A. and Pomranky, Rebecca A. and Pierce, Linda G. and Beck, Heather P.},
  journal={Journal of Experimental Psychology: Applied},
  volume={9},
  number={2},
  pages={86--99},
  year={2003},
  doi={10.1037/1076-898X.9.2.86}
}

@article{madhavan2007trust,
  title={Similarities and Differences Between Human-Human and Human-Automation Trust: An Integrative Review},
  author={Madhavan, Poornima and Wiegmann, Douglas A.},
  journal={Theoretical Issues in Ergonomics Science},
  volume={8},
  number={4},
  pages={277--301},
  year={2007},
  doi={10.1080/14639220500337708}
}

@article{hancock2023how,
  title={How and why humans trust: A meta-analysis and elaborated model},
  author={Hancock, Peter A and Kessler, Theresa T and Kaplan, Alexandra D and Stowers, Kimberly and Brill, J Christopher and Billings, Deborah R and Schaefer, Kristin E and Szalma, James L},
  journal={Frontiers in psychology},
  volume={14},
  pages={1081086},
  year={2023},
  publisher={Frontiers Media SA}
}

@article{legler2025human,
  title={How to achieve human-centered automation: the importance of trust for safety-critical behavior and intention to use in human-robot collaboration},
  author={Legler, Franziska and Bullinger, Angelika C},
  journal={Frontiers in Organizational Psychology},
  volume={3},
  pages={1669782},
  year={2025},
  publisher={Frontiers Media SA}
}

@inproceedings{lightman2023lets,
  title={Let's verify step by step},
  author={Lightman, Hunter and Kosaraju, Vineet and Burda, Yuri and Edwards, Harrison and Baker, Bowen and Lee, Teddy and Leike, Jan and Schulman, John and Sutskever, Ilya and Cobbe, Karl},
  booktitle={The twelfth international conference on learning representations},
  year={2023}
}

@article{uesato2022solving,
  title={Solving math word problems with process-and outcome-based feedback},
  author={Uesato, Jonathan and Kushman, Nate and Kumar, Ramana and Song, Francis and Siegel, Noah and Wang, Lisa and Creswell, Antonia and Irving, Geoffrey and Higgins, Irina},
  journal={arXiv preprint arXiv:2211.14275},
  year={2022}
}

@inproceedings{wu2023autogen,
  title={AutoGen: Enabling Next-Gen LLM Applications via Multi-Agent Conversation},
  author={Wu, Qingyun and others},
  booktitle={The 2023 Conference on Empirical Methods in Natural Language Processing},
  year={2023}
}

@inproceedings{hong2024metagpt,
  title={MetaGPT: Meta programming for a multi-agent collaborative framework},
  author={Hong, Sirui and Zhuge, Mingchen and Chen, Jonathan and Zheng, Xiawu and Cheng, Yuheng and Wang, Jinlin and Zhang, Ceyao and Wang, Zili and Yau, Steven Ka Shing and Lin, Zijuan and others},
  booktitle={The twelfth international conference on learning representations},
  year={2023}
}

@article{schuster2022confident,
  title={Confident adaptive language modeling},
  author={Schuster, Tal and Fisch, Adam and Gupta, Jai and Dehghani, Mostafa and Bahri, Dara and Tran, Vinh and Tay, Yi and Metzler, Donald},
  journal={Advances in Neural Information Processing Systems},
  volume={35},
  pages={17456--17472},
  year={2022}
}

@article{kaplan2020scaling,
  title={Scaling laws for neural language models},
  author={Kaplan, Jared and McCandlish, Sam and Henighan, Tom and Brown, Tom B and Chess, Benjamin and Child, Rewon and Gray, Scott and Radford, Alec and Wu, Jeffrey and Amodei, Dario},
  journal={arXiv preprint arXiv:2001.08361},
  year={2020}
}

\end{document}